\documentclass[letterpaper, 10 pt, conference]{ieeeconf}  

\IEEEoverridecommandlockouts                              

\usepackage{color}
\usepackage[dvipsnames]{xcolor} 
\usepackage{colortbl}
\usepackage{hyperref}
\usepackage{bm}
\usepackage{url}
\usepackage{graphicx}
\usepackage{lipsum}
\usepackage{pifont}
\usepackage{amsmath}
\usepackage{amssymb}
\usepackage{nccmath}
\usepackage{comment}
\usepackage{balance}
\usepackage{multirow}
\usepackage{textcomp}
\usepackage{multirow}
\usepackage{subcaption}
\usepackage{booktabs}
\usepackage{algorithm}
\usepackage{dblfloatfix}
\usepackage[noend]{algpseudocode}

\definecolor{honghao}{RGB}{128, 0, 128}

\usepackage{cite}

\title{\LARGE \bf
SuperLoc: The Key to Robust LiDAR-Inertial Localization Lies in Predicting Alignment Risks \\
\bf \Large \color{orange}{\href{https://superodometry.com/superloc}{superodometry.com/superloc}
}
}

\author{
    Shibo Zhao$^{\dag}$$^{1}$, Honghao Zhu$^{\dag}$$^{2}$, Yuanjun Gao$^{1}$, 
    Beomsoo Kim$^{1}$, Yuheng Qiu$^{1}$, \\
    Aaron M. Johnson$^{2}$, and Sebastian Scherer$^{1}$ \\
    {$^{\dag}$Equal contribution. Corresponding author: shiboz@andrew.cmu.edu} \\
    {$^{1}$The Robotics Institute, $^{2}$Department of Mechanical Engineering, Carnegie Mellon University} \\
}

\begin{document}
\setlength {\marginparwidth }{1cm} 
\twocolumn[{%
    \renewcommand\twocolumn[1][]{#1}%
    \begin{center} 
    \centering 
    \maketitle
    \includegraphics[width=1.0\linewidth]{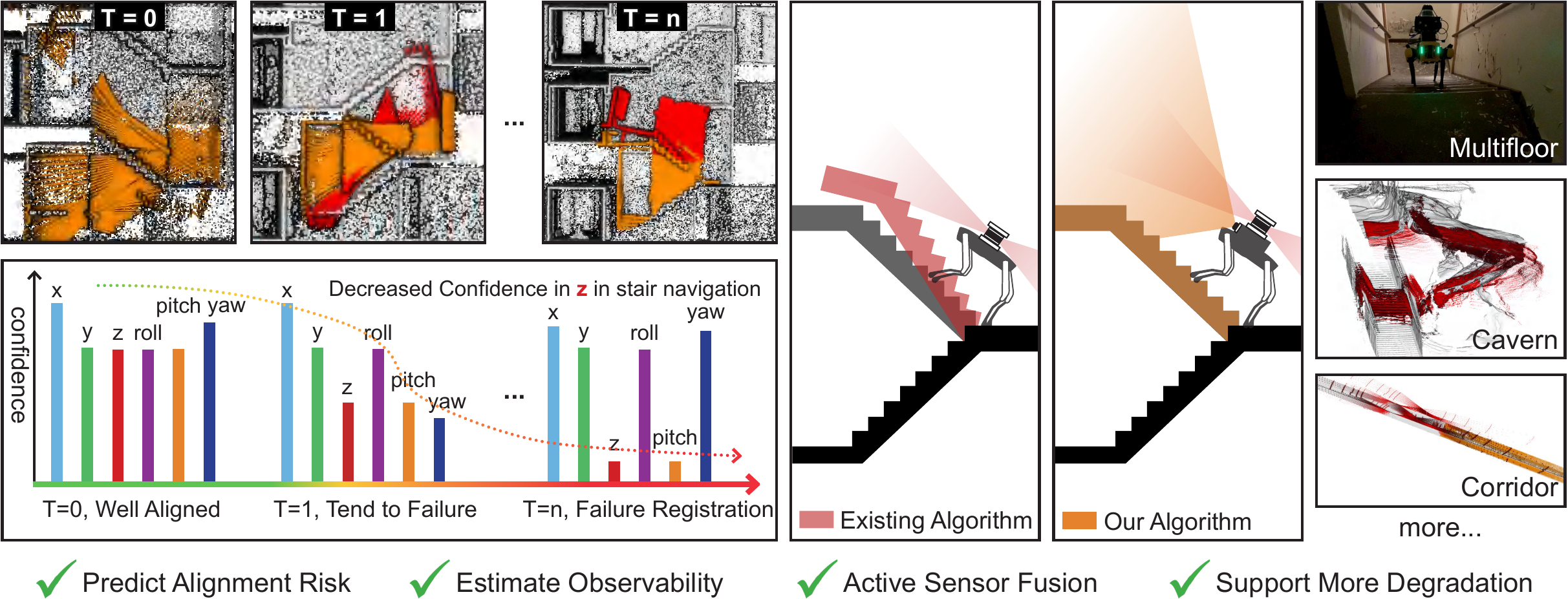}
    \captionof{figure}{SuperLoc is an open-source LiDAR-inertial localization system that not only predicts alignment risks, and estimates the observability of scan, but also can actively incorporate pose priors from other odometry sources before failure occurs. The entire process doesn't require heuristic threshold adjustment to detect degeneration, and it has been evaluated in various challenging environments, including caves, long corridors, flat open areas, and staircases.} 
    \label{fig:first_figure}
    \end{center}%
}]

  

\definecolor{Gray}{gray}{0.90}
\newcolumntype{g}{>{\columncolor{Gray}}c}
\definecolor{ffe1da}{RGB}{255,225,218}
\definecolor{F7E0D5}{RGB}{247,224,213}
\definecolor{darkF7E0D5}{RGB}{209,154,128}

\definecolor{Dark}{rgb}{0,0,0}

\definecolor{OutdoorDark}{rgb}{0,.5,0}
\definecolor{comment}{rgb}{0.6, 0.4, 0.8}
\definecolor{IndoorDark}{rgb}{0,0.3,0.8}
\definecolor{SubTDark}{rgb}{0.5,.27,0.11}
\definecolor{AerialDark}{rgb}{.5,.0,.5}
\definecolor{UnderWaterDark}{rgb}{0.16, 0.46, 0.81}
\colorlet{Outdoor}{OutdoorDark!20!white}
\colorlet{Indoor}{IndoorDark!20!white}
\colorlet{SubT}{SubTDark!20!white}
\colorlet{Aerial}{AerialDark!20!white}
\colorlet{UnderWater}{UnderWaterDark!20!white}
\colorlet{OutdoorLight}{OutdoorDark!70!white}
\colorlet{IndoorLight}{IndoorDark!70!white}
\colorlet{SubTLight}{SubTDark!70!white}
\colorlet{AerialLight}{AerialDark!70!white}
\colorlet{UnderWaterLight}{UnderWaterDark!70!white}

\definecolor{tabfirst}{rgb}{1, 0.85, 0.7}

\newcommand{\symbolHt}{1.5em}
\newcommand{\outdoorChar}{%
  \begingroup\normalfont
  \includegraphics[height=\symbolHt]{figure/symbols/outdoor.png}%
  \endgroup
}
\newcommand{\indoorChar}{%
  \begingroup\normalfont
  \includegraphics[height=\symbolHt]{figure/symbols/indoor.png}%
  \endgroup
}
\newcommand{\subtChar}{%
  \begingroup\normalfont
  \includegraphics[height=\symbolHt]{figure/symbols/cave.png}%
  \endgroup
}
\newcommand{\hawkinsChar}{%
  \begingroup\normalfont
  \includegraphics[height=\symbolHt]{figure/symbols/degraded.png}%
  \endgroup
}
\newcommand{\aerialChar}{%
  \begingroup\normalfont
  \includegraphics[height=\symbolHt]{figure/symbols/aerial.png}%
  \endgroup
}
\newcommand{\underwaterChar}{%
  \begingroup\normalfont
  \includegraphics[height=\symbolHt]{figure/symbols/underwater.png}%
  \endgroup
}
\newcommand{\viewpointChar}{%
  \begingroup\normalfont
  \includegraphics[height=\symbolHt]{figure/symbols/viewpoint.png}%
  \endgroup
}
\newcommand{\lightingChar}{%
  \begingroup\normalfont
  \includegraphics[height=\symbolHt]{figure/symbols/day_night.png}%
  \endgroup
}
\newcommand{\oppsymbolHt}{1em}
\newcommand{\oppositeChar}{%
  \begingroup\normalfont
  \includegraphics[height=\oppsymbolHt]{figure/symbols/opp_viewpoint.png}%
  \endgroup
}

\newcommand{\urban}[1]{\textbf{\textcolor{OutdoorDark}{Urban}}}
\newcommand{\indoor}[1]{\textbf{\textcolor{IndoorDark}{Indoor}}}
\newcommand{\aerial}[1]{\textbf{\textcolor{AerialDark}{Aerial}}}
\newcommand{\subt}[1]{\textbf{\textcolor{SubTDark}{SubT}}}
\newcommand{\degraded}[1]{\textbf{\textcolor{SubTDark}{Degraded}}}
\newcommand{\underwater}[1]{\textbf{\textcolor{UnderWaterDark}{Underwater}}}

\newcommand{\indoorarg}[1]{\textbf{\textcolor{IndoorDark}{#1}}}
\newcommand{\urbanarg}[1]{\textbf{\textcolor{OutdoorDark}{#1}}}
\thispagestyle{empty}
\pagestyle{empty}
\setlength\floatsep{8pt}
\setlength\textfloatsep{8pt}

\begin{abstract}
Map-based LiDAR localization, while widely used in autonomous systems, faces significant challenges in degraded environments due to the lack of distinct geometric features. This paper introduces SuperLoc, a robust LiDAR localization package that addresses key limitations in existing methods. SuperLoc features a novel predictive alignment risk assessment technique, enabling early detection and mitigation of potential failures before optimization. This approach significantly improves performance in challenging scenarios such as corridors, tunnels, and caves. Unlike existing degeneracy mitigation algorithms that rely on post-optimization analysis and heuristic thresholds, SuperLoc evaluates the localizability of raw sensor measurements. Experimental results demonstrate significant performance improvements over state-of-the-art methods across various degraded environments. Our approach achieves a 54\% increase in accuracy and exhibits better robustness. To facilitate further research, we release our implementation along with datasets from eight challenging scenarios.

\end{abstract}
    
    \section{Introduction} 
\label{sec:cave}

Map-based LiDAR localization is a widely used technique in autonomous systems, including autonomous driving, drone inspection, and search and rescue operations. The Iterative Closest Point (ICP) algorithm\cite{censi07icp,brossard20icp,talbot23icp,zhang24preparedworst,laconte23certify} is the predominant method for estimating a robot's pose on a 3D prior map. ICP aligns LiDAR scans by matching corresponding points or features, determining relative pose transformations, and enabling the creation of detailed scene maps. While LiDAR-based localization and mapping algorithms\cite{zhen19,zou23, vizzo2023kiss,jelavic22open3dslam,ram2020lidar} have seen success, achieving robust localization still remains challenges.
The key challenges are:

\textit{Unreliable Performance in Degraded Scenarios:} 
State-of-the-art LiDAR Localization systems\cite{koide2019portable,xu2021fast} can achieve low drift over long distances but often fail in degraded scenarios such as corridors, hallways, tunnels, caves, or planar areas. These environments lack geometric features, leading to insufficient constraints in the optimization process.

\textit{Too Late to Prevent Registration Failure:} 
Existing degeneracy mitigation algorithms~\cite{zhang2016degeneracy,hinduja2019degeneracy} use solution remapping to update poses along well-constrained directions based on hessian matrix. However, these methods are often too late to prevent imminent failures and suffer from numerical instability because of insufficient constraints during degeneracy.

Motivated by these limitations, we propose SuperLoc, a robust LiDAR localization package by extending our previous odometry system \cite{zhao2021super} to a map-based localization scenario.
The proposed method addresses the above:

\begin{enumerate}
    \item \textbf{Predictive Alignment Risk Assessment:} The main contribution of this work is a method that can not only predict the risk of failed alignment between two point clouds but also estimate the confidence of the current scan during the robot operation without suffering from numerical instability in degeneracy.
    \item \textbf{Active Sensor Fusion:} Rather than passively fusing multiple sensors, we estimate observability in the front-end and integrate it into the sensor fusion. This allows us to actively incorporate pose prior regularization from an alternative odometry source at optimal moments. Our active degeneracy mitigation strategy significantly enhances robustness and accuracy, demonstrating a 54\% improvement over existing methods.


    
    
    \item \textbf{Open-Source Contribution} We release our code along with datasets from 8 challenging environments, including ground-truth maps, to push the limits of localization capabilities in extremely challenging scenarios.

\end{enumerate}
We find that the degeneracy problem is not about reducing outliers; rather, it arises from insufficient constraints~\cite{tuna23xicp, tuna24informed}. Providing additional constraints before failures occur can significantly improve the robustness of localization algorithms. This indicates the importance of predicting alignment risk or degeneracy before, rather than during or after, optimization.

\section{Related Work} 

\subsection{Degeneracy Mitigation in Scan Registration}
When robots operate in complex and challenging environments, an ideal LiDAR localization algorithm should be fail-safe and failure-aware to mitigate the effect of ill-conditioned optimization problems.
Most existing LiDAR localization methods passively fuse data from all sensor modalities into a single, large optimization problem~\cite{zheng2022fast,lin2022r}. This approach assumes that different sensor modalities provide complementary constraints in the optimization process. However, this tightly coupled strategy often requires substantial computational resources. Moreover, it is usually difficult to balance weight between multi modalities.
Zhang~\cite{zhang2016degeneracy} proposes a solution remapping to only update the pose in well-constrained state directions. Similar methods are used in\cite{hinduja2019degeneracy}\cite{nubert2022learning} to detect degeneracy during optimization. However, such methods often rely on heuristic tuning of eigenvalues. These approaches are difficult to apply with the same set of parameters in different environments. In contrast, our approach can achieve early failure detection and does not need heuristic tuning by evaluating the observability. 
\subsection{Existing LiDAR Localization Methods}
The open-source community offers several excellent LIO SLAM algorithms such as Fast-LIO\cite{xu2021fast} and LIO-SAM\cite{shan2020lio}, as well as LVIO algorithms like R3LIVE\cite{lin2022r} and FAST-LIVO~\cite{zheng2022fast}. However, map-based localization algorithms remain relatively scarce, with even fewer performing well in challenging environments. Recent advancements include PALoc\cite{hu2024paloc}, which proposes a flexible prior-assisted localization approach using a global factor graph, and HDL\_LOC\cite{koide2019portable}, which introduces a portable measurement system for long-term, wide-area localization on prior maps. Additionally, other localization algorithms~\cite{koide2024tightly,chen2023dliom} have demonstrated high-accuracy pose estimation in large-scale environments. Despite these developments, these methods are only tested in standard conditions and prioritize accuracy over robustness.



\section{Methodology} 
\begin{figure*}[tb]
  \centering
  \includegraphics[width=.85\linewidth]{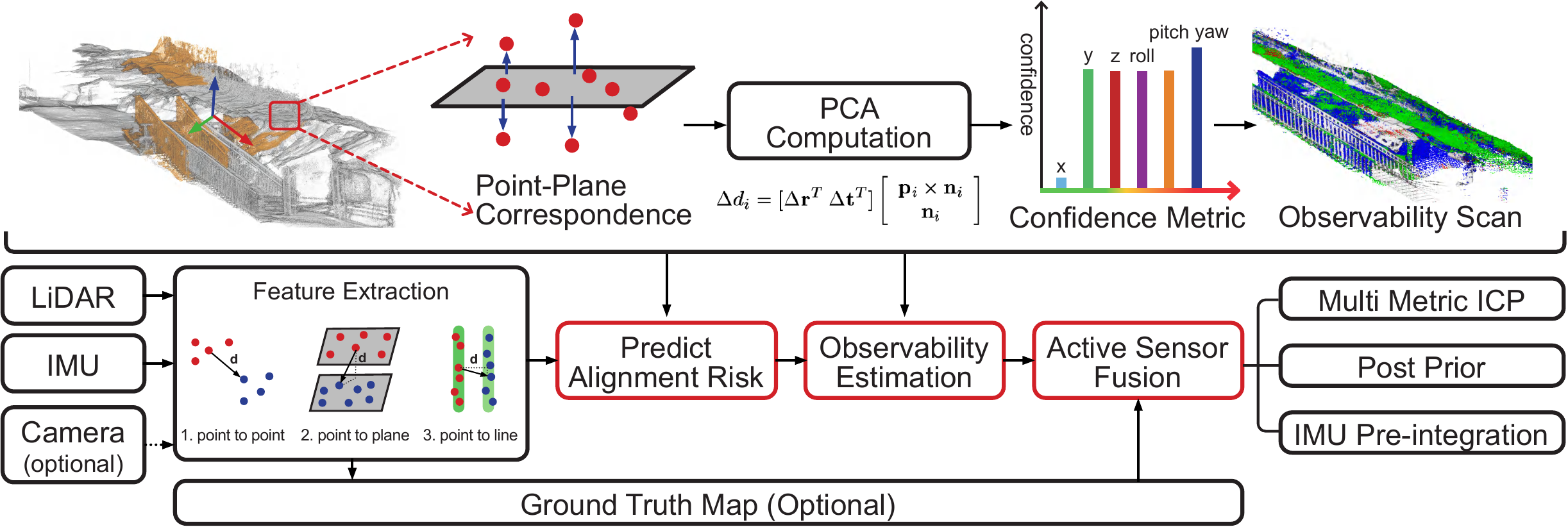}
  \caption{\textbf{System Overview of Proposed Method.} The major contributions of this work are highlighted with a red square. The pipeline begins with LiDAR and IMU measurements, from which point, plane, and line features are extracted. Point-plane correspondences are used for PCA to determine principal and normal directions, which are analyzed for observability. Confidence values of each state direction are used to generate an observability scan, incorporating the pose priors into the sensor fusion system to prevent degeneracy. }
  \label{fig:method}
  \vspace{-0.6 cm}
\end{figure*}


Most sensor fusion methods \cite{xu2021fast,wen24liver,zheng24trajlio} rely on the eigenvalues of the Hessian matrix to quantify state estimation uncertainty, assuming a Gaussian distribution even in degenerate cases. However, this assumption breaks down when certain states or landmarks become unobservable. This approach has three key limitations: \textit{(i)} eigenvalue analysis occurs too late in the optimization process to provide timely remedy; \textit{(ii)} different environments require different thresholds to determine degradation, reducing generalizability; and \textit{(iii)} Hessian-based analysis may suffer from numerical instability, compromising reliability. \textit{(iv)} Difficult to balance the weight between multi modalities.
Here we mainly explain the predicted alignment risk, observability estimation, and active sensor fusion module. Further details can be found in \cite{zhao2021super}.
The overview of our system is shown in  Fig.~\ref{fig:method}.  

\subsection{Predicting Alignment Risk}
To address the limitation (i), we need to detect alignment risk before odometry failure occurs. 
Here, we explain how to quantify the alignment risk when registering two point clouds $P$ and $Q$ using point-plane correspondences. We employ a KD-tree to identify $k$ point-pairs $(\mathbf{p}_i, \mathbf{q}_i)$, where each $\mathbf{q}_i$ has an associated normal vector $\mathbf{n}_i$. 
Our objective is to find a rigid-body transformation {(rotation $\mathbf{R}_l \in \mathrm{SO}(3)$ and a translation $\mathbf{t}_l\in \mathbb{R}^3$ in LiDAR frame $l$)} that minimizes the sum of squared distances from each $\mathbf{p}_i$ to the plane tangent to $Q$ at $\mathbf{q}_i$. The alignment error is:\begin{align}
E &= \sum_{i=1}^{k} \left( (\mathbf{R}_l \mathbf{p}_i + \mathbf{t}_l - \mathbf{q}_i) \cdot \mathbf{n}_i \right)^2
\label{eq:icp0}
\end{align}

For small rotations, we approximate $\mathbf{R}_l$ using the axis-angle representation:
$\mathbf{R}_l \approx \mathbf{I} + [r_l]_{\times}$, where $r_l$ is a small 3D rotation vector, and $[r_l]_{\times}$ is its corresponding skew-symmetric matrix. Substituting this into the Eq.\ref{eq:icp0}:
\begin{align}
\mathbf{R_l} \mathbf{p_i} \approx ( \mathbf{I} + [r_l]_{\times})\mathbf{p_i} = \mathbf{p_i} + \mathbf{r_l} \times \mathbf{p_i}
\label{eq:icp1}
\end{align}

Thus, the error expression simplifies to
\begin{align}E &= \sum_{i=1}^{k} \left( \left( (\mathbf{p}_i - \mathbf{q}_i) \cdot \mathbf{n}_i \right) + \mathbf{r}_l \cdot (\mathbf{p}_i \times \mathbf{n}_i) + \mathbf{t}_l \cdot \mathbf{n}_i \right)^2 
\label{eq:icp2}
\end{align}
Rewriting the Eq.\ref{eq:icp2}, the residual error $\Delta d_i$ for each point-plane pair is:

\begin{equation}
\Delta d_i = \mathbf{J}_i \Delta x=
\begin{bmatrix}
\Delta \mathbf{r}_l^T & \Delta \mathbf{t}_l^T
\end{bmatrix}
\begin{bmatrix}
\mathbf{p}_i \times \mathbf{n}_i \\
\mathbf{n}_i
\end{bmatrix}
\label{eq:icp3}
\end{equation}

Here, the transformation vector $\Delta x=[\Delta \mathbf{r}_l^T , \Delta \mathbf{t}_l^T]$ and $J_{i}$ represents the Jacobian matrix. This formulation reveals that points where $\mathbf{n}_i$ is perpendicular to $\Delta\mathbf{t}_l$, or $\mathbf{p}_i \times \mathbf{n}_i$ is perpendicular to $\Delta\mathbf{r}_l$, do not contribute to the error $E$ \cite{gelfand2003geometrically}. 

Eq.\ref{eq:icp3} is crucial. Since it shows that knowing only the normal vector \(\mathbf{n}_i\) and 3D point \(\mathbf{p}_i\), we can quickly assess constraint observability in each pose direction (\(\mathbf{X, Y, Z}\)) \textit{before the ICP optimization happens}. Analyzing all constraints in a scan, we provide a comprehensive observability assessment to predict alignment risk shown as matrix \(\mathbf{C}_l\) \cite{gelfand2003geometrically}. 

\begin{equation}
\mathbf{C}_l  =
\begin{bmatrix}
\mathbf{p}_1 \times \mathbf{n}_1 & \cdots & \mathbf{p}_k \times \mathbf{n}_k \\
\mathbf{n}_1 & \cdots & \mathbf{n}_k
\end{bmatrix}
\begin{bmatrix}
(\mathbf{p}_1 \times \mathbf{n}_1)^T & \mathbf{n}_1^T \\
\vdots & \vdots \\
(\mathbf{p}_k \times \mathbf{n}_k)^T & \mathbf{n}_k^T
\end{bmatrix}
\label{eq:alignment_matrix}
\end{equation}

To express our constraints in the world coordinate, we need to apply rotation transformation $\mathbf{C}_w = \mathbf{R}_w \mathbf{C}_l \mathbf{R}_w^T$
\begin{equation}
    \mathbf{C}_w = \sum_{i=1}^{k} 
\begin{bmatrix}
(\mathbf{p}^w_i \times \mathbf{n}^w_i)(\mathbf{p}^w_i \times \mathbf{n}^w_i)^T & (\mathbf{p}^w_i \times \mathbf{n}^w_i)(\mathbf{n}^w_i)^T \\
\mathbf{n}^w_i(\mathbf{p}^w_i \times \mathbf{n}^w_i)^T & \mathbf{n}^w_i (\mathbf{n}^w_i)^T
\end{bmatrix}
\end{equation}

$\mathbf{p}^w_i,\mathbf{n}^w_i$ is point and normal vector in the world frame. 




\subsection{Observability Estimation for State Estimation}
To address the limitations (ii) and (iii), we need to create a more reliable covariance matrix for state estimation. Our key insight is that a well-conditioned problem has constraints distributed evenly across all directions, while an ill-conditioned problem has fewer constraints in specific degraded directions. \textit{If we can identify constraint directions, we can statistically count the number of constraints in each direction and predict the unobservable direction.} 

\begin{table}[h]
    \centering
    \begingroup
    \setlength{\arrayrulewidth}{0.8pt} 
    \renewcommand{\arraystretch}{1.5} 
    \setlength{\tabcolsep}{12pt} 
    \begin{tabular}{|c|c|}
        \hline
        \textbf{Motion Direction Label} & \textbf{Num of Motion Label} \\
        \hline
      X Translation & \( \sum_i \mathbf{L}(n_{ix}) \) \\
        \hline
        Y Translation & \( \sum_i \mathbf{L}(n_{iy}) \) \\
        \hline
        Z Translation & \( \sum_i \mathbf{L}(n_{iz}) \) \\
        \hline
        Roll (X-axis rotation) & \( \sum_i \mathbf{L}(y_i^w n_{iz}^w - z_i^w n_{iy}^w) \) \\
        \hline
        Pitch (Y-axis rotation) & \( \sum_i \mathbf{L}(z_i^w n_{ix}^w - x_i^w n_{iz}^w) \) \\
        \hline
        Yaw (Z-axis rotation) & \( \sum_i \mathbf{L}(x_i^w n_{iy}^w - y_i^w n_{ix}^w) \) \\
        \hline
    \end{tabular}
    \endgroup
    \caption{ The Number of Observability Labels $\mathbf{L}$ for Each Motion Direction. The Normal Vector $\mathbf{n}_i = (n_{ix}, n_{iy}, n_{iz})$. }
    \label{tab:observability_conditions}
\end{table}


According to \( \mathbf{C}^w \), we determine the motion observability label (X, Y, Z, Roll, Pitch, Yaw) for each correspondence by selecting the direction with the highest observability. To assess the overall observability of a scan, we statistically analyze the distribution of observability labels across each motion state, as summarized in Table \ref{tab:observability_conditions}.

Let $\mathcal{O} = \{x, y, z, \text{roll}, \text{pitch}, \text{yaw}\}$ be the set of observable dimensions in our state space. For each dimension $i \in \mathcal{O}$, we define $N_i$ as the count of observability labels in that dimension.
The total observability count is given by $N_{\text{total}} = \sum_{i \in \mathcal{O}} N_i$. We define the normalized confidence metrics as:

\begin{align}
\boldsymbol{\gamma}_{\text{trans}} &= {|\mathcal{O}|} \begin{bmatrix}
\frac{N_x}{N_{\text{total}}}, & 
\frac{N_y}{N_{\text{total}}}, & 
\frac{N_z}{N_{\text{total}}}
\end{bmatrix}^T \\
\boldsymbol{\gamma}_{\text{rot}} &= {|\mathcal{O}|} \begin{bmatrix}
\frac{N_{\text{roll}}}{N_{\text{total}}}, & 
\frac{N_{\text{pitch}}}{N_{\text{total}}}, & 
\frac{N_{\text{yaw}}}{N_{\text{total}}}
\end{bmatrix}^T
\end{align}

Here, the factor ${|\mathcal{O}|} = 6$ scales the metrics to $[0,1]$ under uniform distribution across all dimensions. 
These confidence metrics constructs a diagonal covariance matrix:

\begin{align}
\boldsymbol{\Sigma}_{cov} = \text{diag}(\boldsymbol{\gamma}_{\text{trans}}, \boldsymbol{\gamma}_{\text{rot}})
\label{eq:cov}
\end{align}
 It’s important to note that
the Eq.\ref{eq:cov} assesses the ratio of observability labels in
a specific direction to the total number of labels. This
relative approach is based on the hypothesis that in well-
structured environments, observability should be uniformly
distributed across all state directions. Consequently, our
proposed method eliminate the need for heuristic threshold adjustment since it is a relative value ranges from [0,1]. We evaluated the covariance matrix, denoted as $\boldsymbol{\Sigma}_{cov}$, across various environments. We found that when any element of $\boldsymbol{\Sigma}_{cov}$ is less than 0.2, there is a high likelihood of degradation in that direction. In most cases, this threshold typically does not require adjustment for different scenes.

\subsection{Active Sensor Fusion}    
Traditional sensor fusion approaches passively combine all measurements and rely on robust kernel functions to mitigate adverse effects\cite{chebrolu2021adaptive}. To address the limitation (iv), We propose a more proactive strategy: degeneracy-aware sensor fusion. This actively incorporates additional constraints with proper weighting, particularly in degenerate directions. Our method not only identifies degraded directions but also determines the necessary weight in those directions using the previous proposed $\boldsymbol{\Sigma}_{cov}$ matrix. 


\subsubsection{Pose Prior Factor}
When degeneracy occurs, we leverage relative poses from alternative odometry sources, denoted as [$\Delta p_j^i$, $\Delta q_j^i$] between two consecutive frames $i$ and $j$ in local coordinates. These are used to constrain the relative position $\hat \alpha _j^i$ and rotation $\hat \gamma _j^i$ estimates obtained from joint optimization. We define the covariance matrix of this prior information as $\mathbf{Cov_{prior}}=\mathbf{I_{6x6}}-\boldsymbol{\Sigma}_{cov}$. This regularization term ensures that constraints are equally distributed across state directions to mitigate the effects of numerical instability.
The relative pose factor $\mathbf{e}_{i j}$ is defined:

\begin{equation}
\mathbf{e}_{i j}^{prior}=\left[ {\begin{array}{*{20}{c}}
{\Delta p_j^i}\\
{\Delta q_j^i}
\end{array}} \right] \ominus \left[ {\begin{array}{*{20}{c}}
{\hat \alpha _j^i}\\
{\hat \gamma _j^i}
\end{array}} \right]
\end{equation}
Notably, in environments with well-distributed constraints, the pose prior carries minimal weight.
\subsubsection{Joint Optimization}

We employ a factor graph approach to minimize all residual terms:

\begin{equation}
\mathop {\min }\limits_{{\mathbf{T}_{i + 1}}} \left\{ {\left. {\begin{array}{*{20}{c}}
{\sum\limits_{p \in {\mathbb{F}_{i}}} {{{\left\| {e_i^{po \to po,li,pl}} \right\|}_{\mathbf{cov}}^2} + } }\\
{\sum\limits_{\left( {i,i + 1} \right) \in \mathcal{B} } {{{\left\| {{e^{imu}}} \right\|}^2} }+\sum\limits_{\left( {i,i + 1} \right) \in \mathcal{B} } {{{\left\| {{e^{prior}}} \right\|}_{\mathbf{cov_{prior}}}^2}}}
\end{array}} \right\}} \right.
\label{eq:pgo_eqn}
\end{equation}

Here, $e^{prior}$ is the relative pose factor, actively fused and weighted by the proposed $\mathbf{Cov_{prior}}$ and $e_i^{po \to po,li,pl}$ represents point-to-point (line, distance) residuals between two frames\cite{zhao2021super}, $e^{imu}$ denotes the IMU preintegration factor\cite{Forster_2015}. For more details, please refer to \cite{zhao2021super}.

    \section{Experiment} 
In this section, we assess the viability of our proposed approach and evaluate its performance in real-world scenarios. Our analysis consists of three main stages:

\begin{itemize}
    \item \textbf{Alignment Risks Prediction:} We assess our method on potential failure scans (caves, tunnels, corridors) and validate our alignment risk prediction using our covariance matrix.
    \item \textbf{Active Sensor Fusion:}  Leveraging covariance analysis, we integrate prior odometry to mitigate degeneracy and demonstrate improved map accuracy and robustness.
    \item \textbf{Odometry Evaluation:} We thoroughly assess the accuracy and robustness of our approach in challenging environments. 
\end{itemize}

\label{sec:cave}
\begin{figure}[tb]
  \centering
  \includegraphics[width=0.8\linewidth]{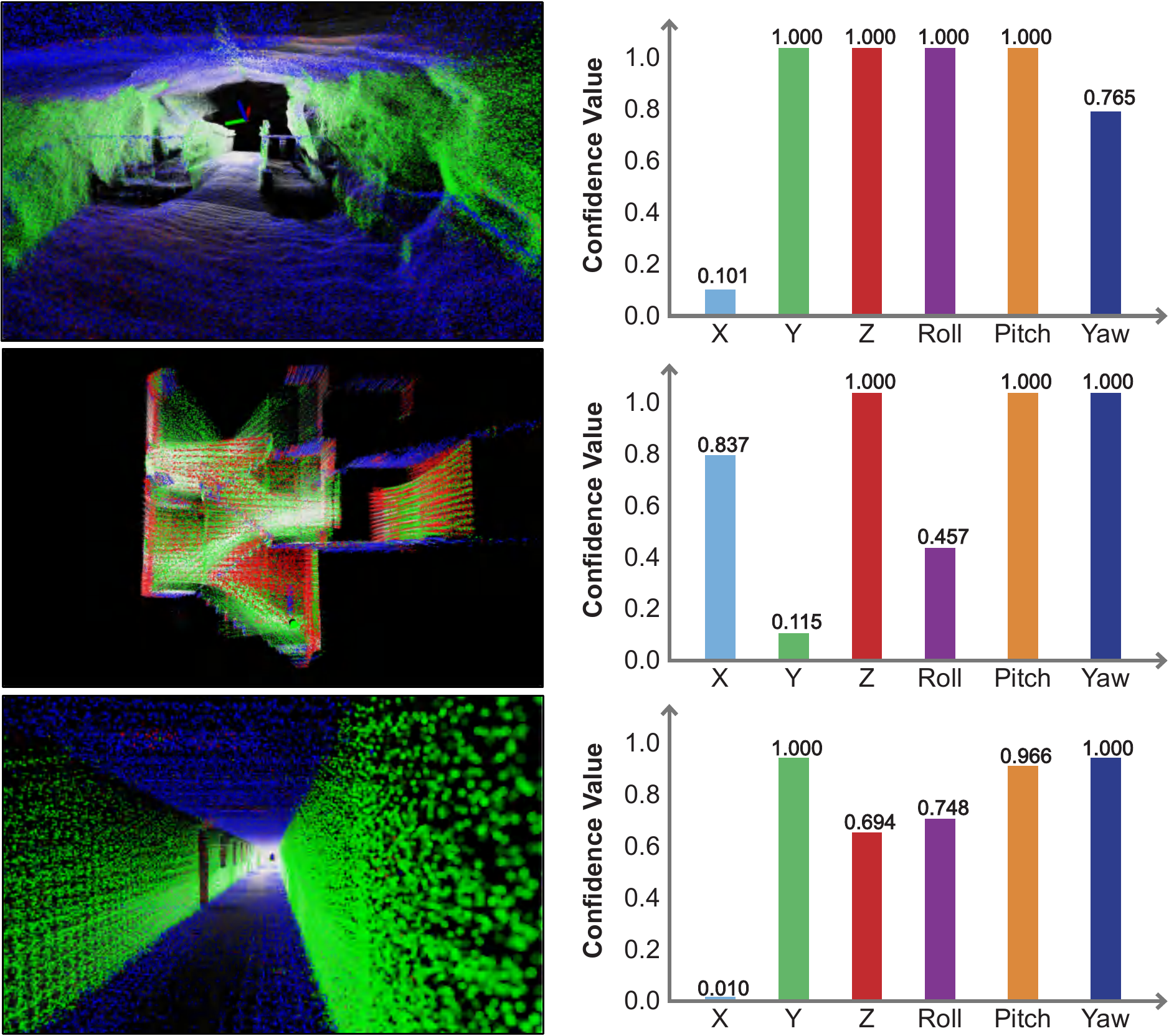}
  \caption{\textbf{Real-Time Alignment Risk Analysis.} From top to bottom: Cave, Multi-floor, and Long-corridor environments. Left: Observability scans, where red, green, and blue points represent accumulated observable features in the x, y, and z directions, respectively. Right: Histograms depicting real-time confidence values $[0, 1]$ for each direction. Lower confidence values indicate higher alignment risks.}
  \label{fig:obser}
\end{figure}
\subsection{Alignment Risks Prediction}
To evaluate the effectiveness of our approach, we conducted analyses across three challenging environments: caves, multi-floor structures, and long corridors (Fig.~\ref{fig:first_figure}). For each setting, we aggregated approximately 10 seconds of point cloud data and projected our confidence metric onto the 3D space, creating an observability scan.
In both cave and long-corridor environments, our platform's forward movement revealed significantly lower confidence values in the X (forward) direction compared to other dimensions. This phenomenon is visually represented in the observability scans (Fig. \ref{fig:obser}), where the scarcity of red points indicates reduced correspondence in the X direction.
The observability scans provide a clear visual representation of degeneracy in specific directions: a lack of red points (X direction) for caves and long corridors, and fewer blue points (Z direction) for multi-floor environments.


\label{sec:cave}
\subsection{Active Sensor Fusion} 
\begin{figure}[tb]
  \centering
  \includegraphics[width=0.90\linewidth]{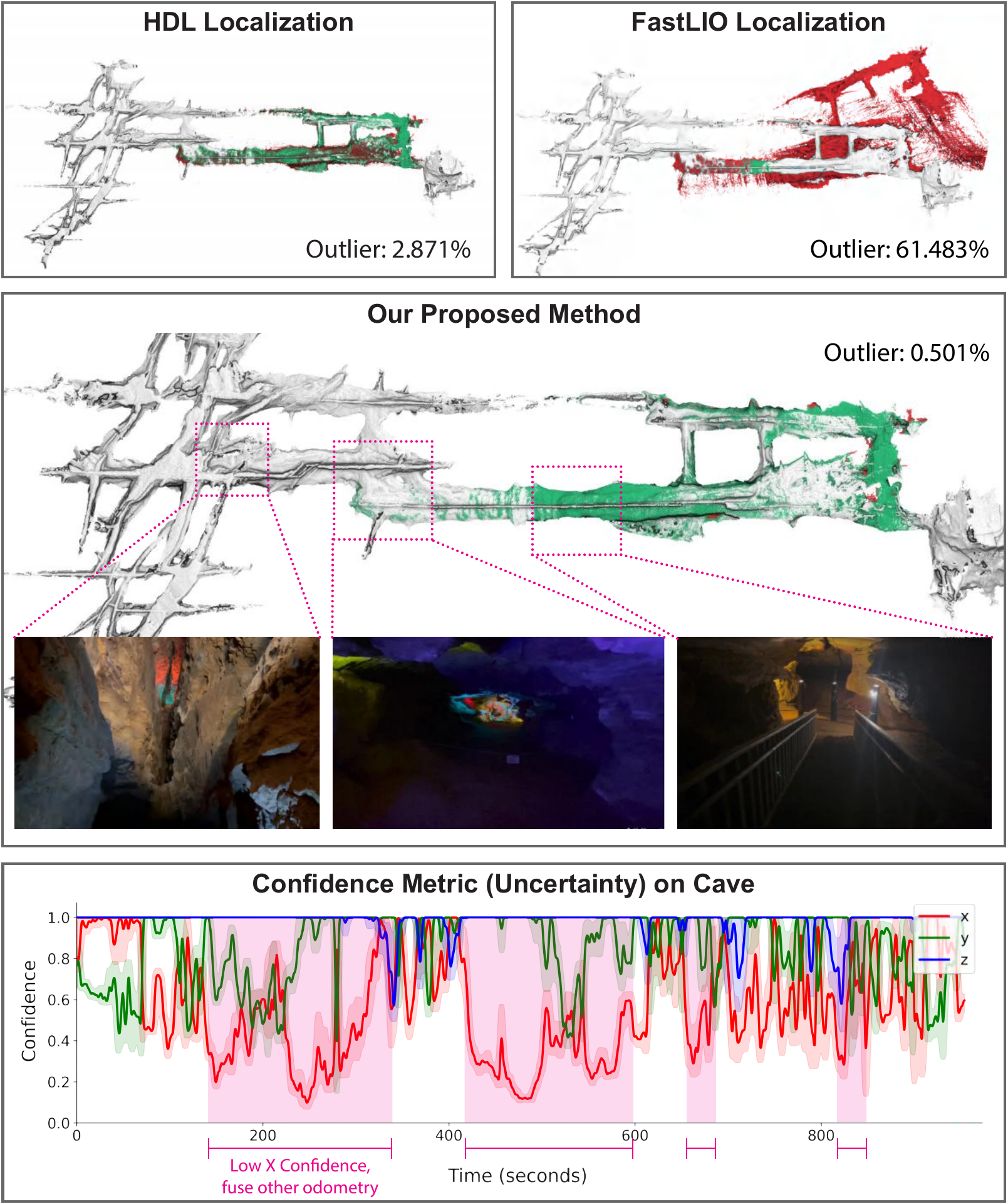}
  \caption{\textbf{Localization in Cave Environments}. Our method demonstrates superior performance with a significantly lower outlier rate of 0.50\%, marking a 6-fold times improvement over the
second-best method.   The pink highlights at the bottom shows low confidence in the X direction, where our method actively fuses alternative odometry in X direction.}
  \label{fig:cave}
\end{figure}


In this section, we demonstrate how to leverage confidence estimation in a timely way to incorporate supplementary odometry to enhance localization robustness in challenging environments. We conducted field experiments using handheld devices, legged robots, and ground vehicles in three degraded environments:
i)   narrow, self-similar cave environments,
ii)  confined, multi-floor environments, and
iii) long corridors with flat open spaces.
To assess the accuracy of our localization algorithm, we employed a FARO scanner to establish a ground truth map with a precision error of is less than 2mm. For each experimental run, we conducted map analysis to quantify the proportion of inliers and outliers in the localization-generated map relative to the ground truth map. We defined inliers and outliers using a 10 cm threshold. In each scenario, we compared our algorithm with HDL Localization \cite{koide2019portable} and FASTLIO Localization \cite{xu2021fast}. We provided ground truth map and initial pose for all the localization algorithms.

\subsubsection{\textbf{Cave Experiments}}
\newcommand{\cmark}{\ding{51}}%
\newcommand{\xmark}{\ding{55}}%
\setlength{\tabcolsep}{5pt}

We evaluated our localization algorithm in Laurel Caverns, a challenging cave environment located in the suburbs of Pittsburgh, USA, as illustrated in Fig. \ref{fig:cave}. This site poses significant challenges for LiDAR SLAM systems due to its smooth, repetitive structures and poor lighting conditions.
Despite these challenges, our proposed method demonstrated superior performance compared to prior approaches across multiple runs, as shown in Table \ref{tab:results_combined}. For Cave01, it achieved an outlier rate of only 0.50\% over 416 meters of travel, representing a 6-fold improvement over the second-best method (2.87\% on the same run). This exceptional performance can be attributed to two key factors:
i) effective prediction of scan registration risks before failures occur; and
ii) accurate estimation of confidence in each state direction, enabling precise fusion of alternative odometry. As shown in Fig. \ref{fig:cave}, our method addresses LiDAR SLAM's X-directional performance degradation by incorporating additional odometry data with appropriate confidence values.


\subsubsection{\textbf{Multi-Floor Environments}} 
\begin{figure}[tb]
  \centering
  \includegraphics[width=0.90\linewidth]{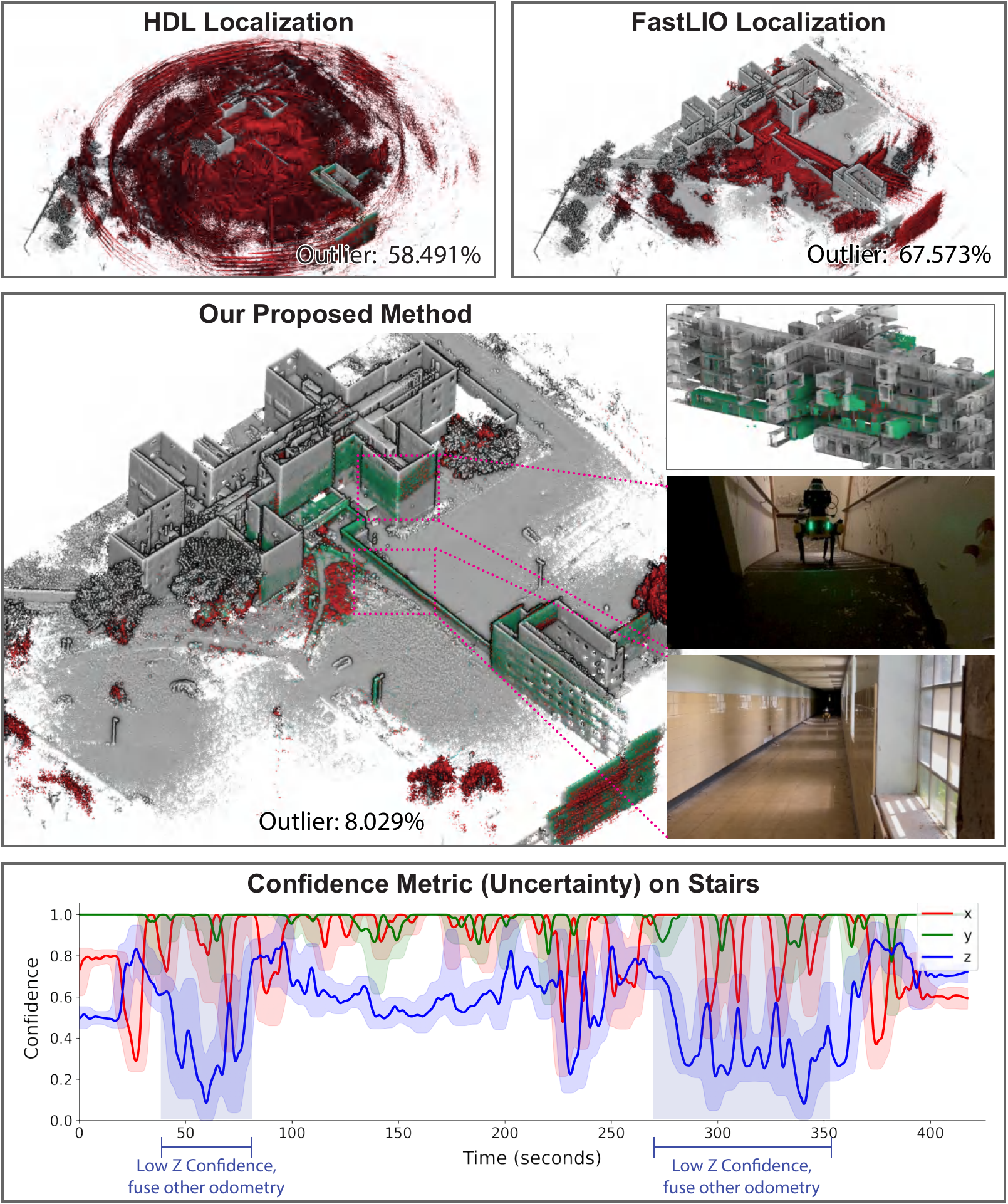}
  \caption{\textbf{Localization in MultiFloor Environments}.  Our method has significantly fewer outliers, with a rate of 8.03\%, marking a 7.2-fold times improvement over the second-best method. The blue dashed square at the bottom highlights low confidence in the Z direction, where our method actively fuses alternative odometry in Z direction. }
  \label{fig:multi_floor}
\end{figure}

We evaluated our algorithm in multi-floor environments using a legged robot to navigate an abandoned hospital. This scenario presents unique challenges due to confined spaces like staircases, which restrict the sensor's field of view. Moreover, even small errors in elevation estimation can lead to significant misalignments between floors, as shown in Fig. \ref{fig:first_figure}.
Our method demonstrated superior performance in this challenging environment, Table \ref{tab:results_combined}. It achieved an outlier rate of only 8.03\% over 270 meters of travel, representing a 7.2-fold improvement over the second-best method (58.49\%).
Fig. \ref{fig:multi_floor} highlights a key feature of our approach: the blue dashed square at the bottom of the figure indicates an area of low confidence in the Z direction. In this region, our method actively incorporates additional odometry in the Z dimension.


\label{sec:long-corridor}
\subsubsection{\textbf{Long Corridor Experiment}} 
\begin{figure}[tb]
  \centering
  \includegraphics[width=0.88\linewidth]{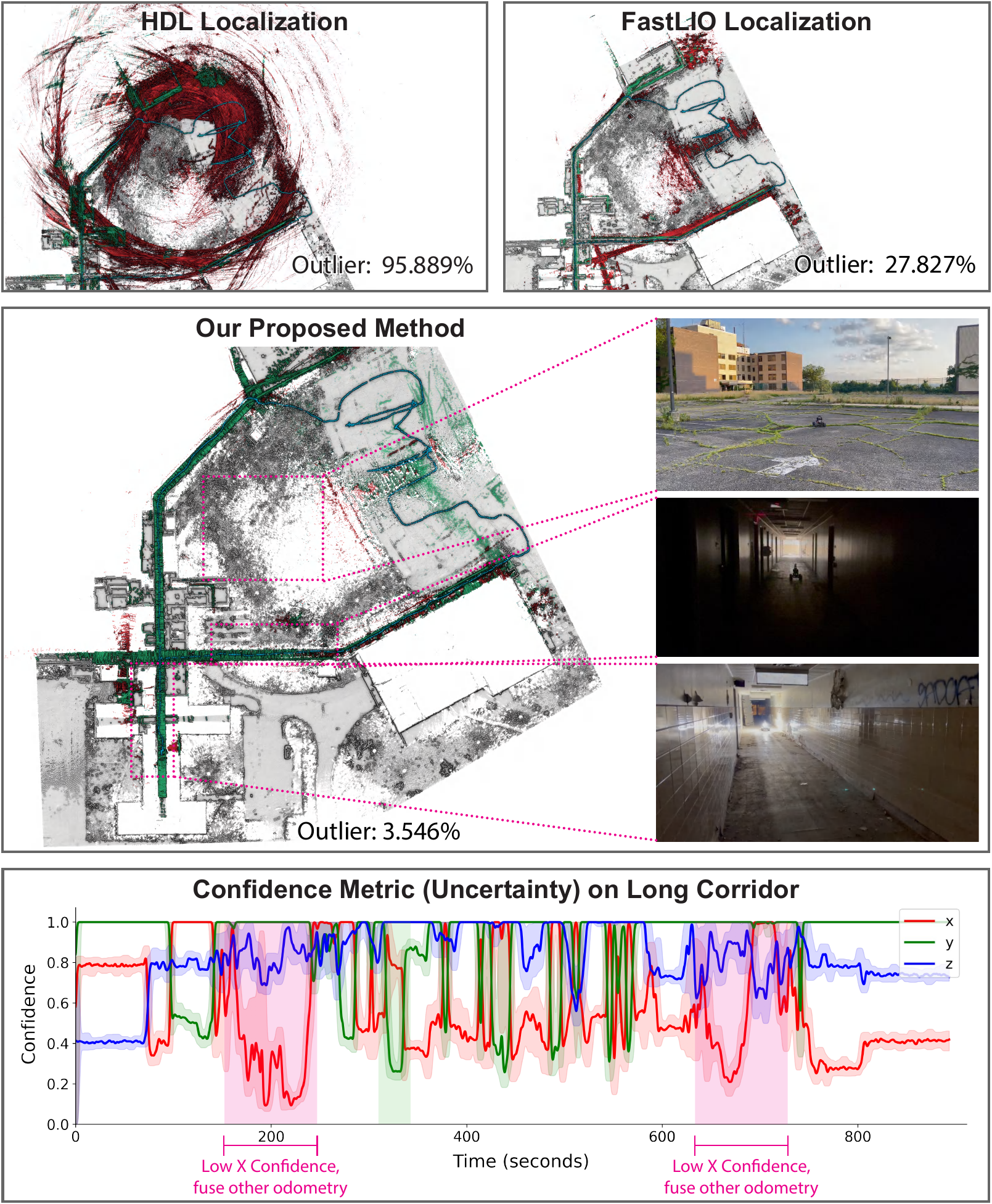}
  \caption{\textbf{Localization in Long Corridor and Open Flat Environments}.  Our method has significantly fewer outliers, with a rate of 3.55\%, marking a 7.8-fold times improvement over the second-best method (27.83\%) over 690 meters of travel. The red, green dashed square at the bottom highlights low confidence in the X and Y direction, where our method actively fuses alternative odometry in those direction}
  \label{fig:long_corridor}
\end{figure}

We used an RC car for the long corridor experiment, which operated at high speeds of up to 5 m/s and 57°/s, navigating both indoor corridors and outdoor open areas. Long corridors present significant challenges for LiDAR systems due to their repetitive, symmetrical structure and lack of distinct geometric features. (Fig. \ref{fig:long_corridor}).
Despite these challenges, our algorithm demonstrated a 7.8-fold improvement over the second-best method, Table \ref{tab:results_combined}. This improvement stems from predicting alignment risks, identifying degradation direction, and assigning appropriate confidence weights during sensor fusion. Notably, our method did not require threshold adjustment to detect point cloud degeneration as the environment changed.


\begin{table}[tb]
  \caption{\centering Outlier Percentages for Cave, Multi-floor, and Long-corridor sequences}
  \label{tab:results_combined}
  \centering
  \scriptsize
  \setlength{\tabcolsep}{4pt}  
  \begin{tabular}{@{}l@{\hspace{6pt}}|@{\hspace{4pt}}c@{\hspace{2pt}}c@{\hspace{2pt}}c@{\hspace{2pt}}c@{\hspace{2pt}}|@{\hspace{2pt}}c|@{\hspace{1pt}}c@{\hspace{1pt}}c@{}}
  \toprule
                    & \multicolumn{7}{c@{}}{\% of Outliers} \\
  Method            & Cave01 & Cave02 & Cave03 & Cave04 & Floor01 & Corridor01 & Corridor02 \\
  Distance          & 416m   & 475m   & 490m   & 597m   & 270m    & 617m   & 690m   \\
  \midrule
  HDL\_LOC          & 2.87 & 52.80 & 2.15 & 4.88 & 58.49 & 27.83 & 42.45 \\
  FAST\_LIO\_LOC    & 61.48 & 25.32 & 13.70 & 41.54 & 67.57 & 95.89 & 73.99 \\
  SuperLoc          & \textbf{0.50} & \textbf{0.81} & \textbf{1.50} & \textbf{1.60} & \textbf{8.03} & \textbf{3.55} & \textbf{23.99} \\
  \bottomrule
  \end{tabular}
  \label{tab:outlier}
\end{table}
\renewcommand{\labelenumi}{\Alph{enumi})}

\begin{table}[tb]
\caption{\centering ATE Performance on SubT-MRS\cite{Zhao2024CVPR}. * denotes incorporation of loop closure. - denotes incomplete trial.}
\scalebox{0.74}{
\begin{tabular}{@{}c@{\hspace{3pt}}|@{\hspace{3pt}}c@{\hspace{3pt}}c@{\hspace{3pt}}c@{\hspace{3pt}}c@{\hspace{3pt}}c@{\hspace{3pt}}|@{\hspace{3pt}}c@{\hspace{3pt}}c@{\hspace{3pt}}c@{\hspace{3pt}}|@{\hspace{3pt}}c@{}}
\toprule
\multirow{2}{*}{\textbf{Method}} & \multicolumn{5}{c|}{\color{Dark} \textbf{Geometric Degradation (SubT-MRS)}} &\multicolumn{3}{c|}{\color{Dark} \textbf{Mix Degradation}} & \multirow{2}{*}{\textbf{Average}} \\
\cmidrule(lr{0.5em}){2-6}  \cmidrule(lr{0.5em}){7-9}
 
 & \color{Dark} Final01  & \color{Dark} Final02 & \color{Dark} Final03 & \color{Dark} Urban01  & \color{Dark} Urban02  & \color{Dark} {Cave03}   & \color{Dark}Corridor01 & \color{Dark}Floor01 & \\ 
\cmidrule(lr){1-1} \cmidrule(lr){2-2} \cmidrule(lr){3-3} \cmidrule(lr){4-4} \cmidrule(lr){5-5} \cmidrule(lr){6-6} \cmidrule(lr){7-7} \cmidrule(lr){8-8} \cmidrule(lr){9-9} \cmidrule(lr){10-10}
Liu\text{*}~\cite{xu2022fast2}& 0.307  & 0.095 & 0.629 & 0.122 & 0.235 & 0.260      & 1.454 & 0.401  & 0.588\\
Weitong\text{*}~\cite{faster_lio}  & 0.26 & 0.096 & 0.617 & 0.120 & 0.222 & 0.402 & 1.254 & 0.577 & 0.663 \\
Kim\text{*}~\cite{xu2022fast2} & 0.331          & 0.092         & 0.787      & 0.123        & 0.270 & 0.279  & 2.100   & 0.650 & 3.825  \\
Yibin ~\cite{vizzo2023kiss}& 1.060          &0.220         & 0.750           & 0.470         & 0.620 & 9.140          & 2.990 & 5.500 & 4.312  \\
Zhong\text{*}~\cite{chen2022direct}   & 1.205 & 0.695 & - & 1.175 & 1.72 & 2.08 & -  & - & 1.209  \\
Zheng~\cite{zheng2022fast}  & - & - & - & - & - & 3.786 & 55.205  & 19.769 & 26.254  \\
Our  & \textbf{0.238} & \textbf{0.074} & \textbf{0.396} & \textbf{0.026} & \textbf{0.104} & \textbf{0.204} & \textbf{0.817} & \textbf{0.315} & \textbf{0.272}  \\
\midrule
\end{tabular}
}
\label{tab:AccLidar}
\end{table}

\begin{table}[t]
\caption{\centering Robustness Performance on SubT-MRS\cite{Zhao2024CVPR}. * denotes loop closure usage. - denotes incomplete trial.}
\vspace{-3pt}
\centering
\scalebox{0.70}{
\begin{tabular}{@{}l@{\hspace{3pt}}|@{\hspace{3pt}}c@{\hspace{3pt}}|@{\hspace{3pt}}c@{\hspace{3pt}}c@{\hspace{3pt}}c@{\hspace{3pt}}c@{\hspace{3pt}}c@{\hspace{3pt}}|@{\hspace{3pt}}c@{\hspace{3pt}}c@{\hspace{3pt}}c@{\hspace{3pt}}|@{\hspace{3pt}}c@{}}
\toprule
\multicolumn{2}{c|}{} &\multicolumn{5}{c|}{\color{Dark} \textbf{Geometric Degradation (SubT-MRS)}} &\multicolumn{3}{c|}{\color{Dark} \textbf{Mix Degradation}}&\\
\cmidrule(lr{0.5em}){3-7}  \cmidrule(lr{0.5em}){8-10} 
 
\multicolumn{2}{l|}{\textbf{Method}} & \color{Dark} Final01  & \color{Dark} Final02 & \color{Dark} Final03 & \color{Dark} Urban01  & \color{Dark} Urban02  & \color{Dark} {Cave03}   & \color{Dark}Corridor01 & \color{Dark}Floor01 & \textbf{Average}\\ 
\cmidrule(lr){1-2}  \cmidrule(lr){3-3} \cmidrule(lr){4-4} \cmidrule(lr){5-5} \cmidrule(lr){6-6} \cmidrule(lr){7-7} \cmidrule(lr){8-8} \cmidrule(lr){9-9} \cmidrule(lr){10-10} \cmidrule(lr){11-11}
Weitong\text{*}\cite{faster_lio}  & \multirow{7}{*}{Rp} & 0.922 & 0.929 & 0.906 & 0.933 & 0.919 & 0.830 & 0.889 & 0.909 & 0.905  \\
Liu\text{*}\cite{xu2022fast2}  && 0.922 & 0.929 & 0.906 & 0.933 & 0.919 & 0.830 & 0.885 & 0.905 & 0.904\\
Kim\text{*}\cite{xu2022fast2} & & 0.884 & 0.929 & 0.906 & 0.933 & 0.915 & 0.830 & 0.890 & 0.252 & 0.817  \\
Yibin\cite{vizzo2023kiss} & & 0.849 & 0.897 & 0.827 & 0.875 & 0.795 & 0.751 & 0.502 & 0.738 & 0.779    \\
Zhong\text{*}~\cite{chen2022direct} & & 0.278 & 0.910 & 0.827 & 0.905 & 0.877 & - & - & - & 0.759 \\
Zheng\cite{zheng2022fast}  & & - & - & - & - & - & 0.618 & 0.579 & 0.389 & 0.529 \\
Our&  &  \textbf{0.929} & \textbf{0.933} & \textbf{0.916} & \textbf{0.937} & \textbf{0.925} & \textbf{0.973} & \textbf{0.899} & \textbf{0.891} & \textbf{0.925}\\
\midrule
Weitong \text{*}\cite{faster_lio} & \multirow{7}{*}{Rr}& 0.931 & 0.931 & 0.907 & 0.938 & 0.924 & 0.848 & 0.890 & 0.914 & 0.911\\
Liu\text{*}\cite{xu2022fast2} && 0.931 & 0.931 & 0.908 & 0.938 & 0.924 & 0.848 & 0.885 & 0.908 & 0.909\\
Kim\text{*}\cite{xu2022fast2} & & 0.923 & 0.931 & 0.908 & 0.938 & 0.920 & 0.848 & 0.890 & 0.260 & 0.827  \\
Yibin\cite{vizzo2023kiss} & & 0.924 & 0.928 & 0.898 & 0.933 & 0.914 & 0.846 & 0.874 & 0.886 & 0.900\\
Zhong\text{*}~\cite{chen2022direct}& & 0.296 & 0.921 & 0.851 & 0.922 & 0.905 & - & - & - & 0.779\\
Zheng\cite{zheng2022fast}  & & - & - & - & - & - & 0.629 & 0.882 & 0.489 & 0.667 \\
Our& &  \textbf{0.938} & \textbf{0.942} & \textbf{0.925} & \textbf{0.943} & \textbf{0.941} & \textbf{0.964} & \textbf{0.944} & \textbf{0.924} & \textbf{0.940}\\
\midrule
\end{tabular}
}
\label{tab:RobustLidar}
\end{table}
\subsection{Odometry Accuracy Evaluation} 
To further validate the accuracy of pose estimation, we conducted Absolute Trajectory Error (ATE) analysis using our odometry system \cite{zhao2021super} on the SubT-MRS dataset \cite{Zhao2024CVPR}. This dataset encompasses challenging environments featuring sensor degradation, aggressive locomotion, and extreme weather conditions. The eight sequences in the dataset are categorized into two groups for testing: Geometric degradation (point cloud degradation only) and Mixed degradation (both point cloud and visual degradation). ATE results for competing systems were obtained from an open SLAM challenge \cite{Zhao2024CVPR}. Participating teams were permitted to employ loop closure techniques for post-processing accuracy enhancement. Despite this, our method still surpasses the second-best method by 54\%, demonstrating an average ATE of 0.271 without employing loop closure (Table~\ref{tab:AccLidar}). 
\subsection{Odometry Robustness Metric Evaluation}
\begin{figure}[tb]
  \centering
  \includegraphics[width=\linewidth]{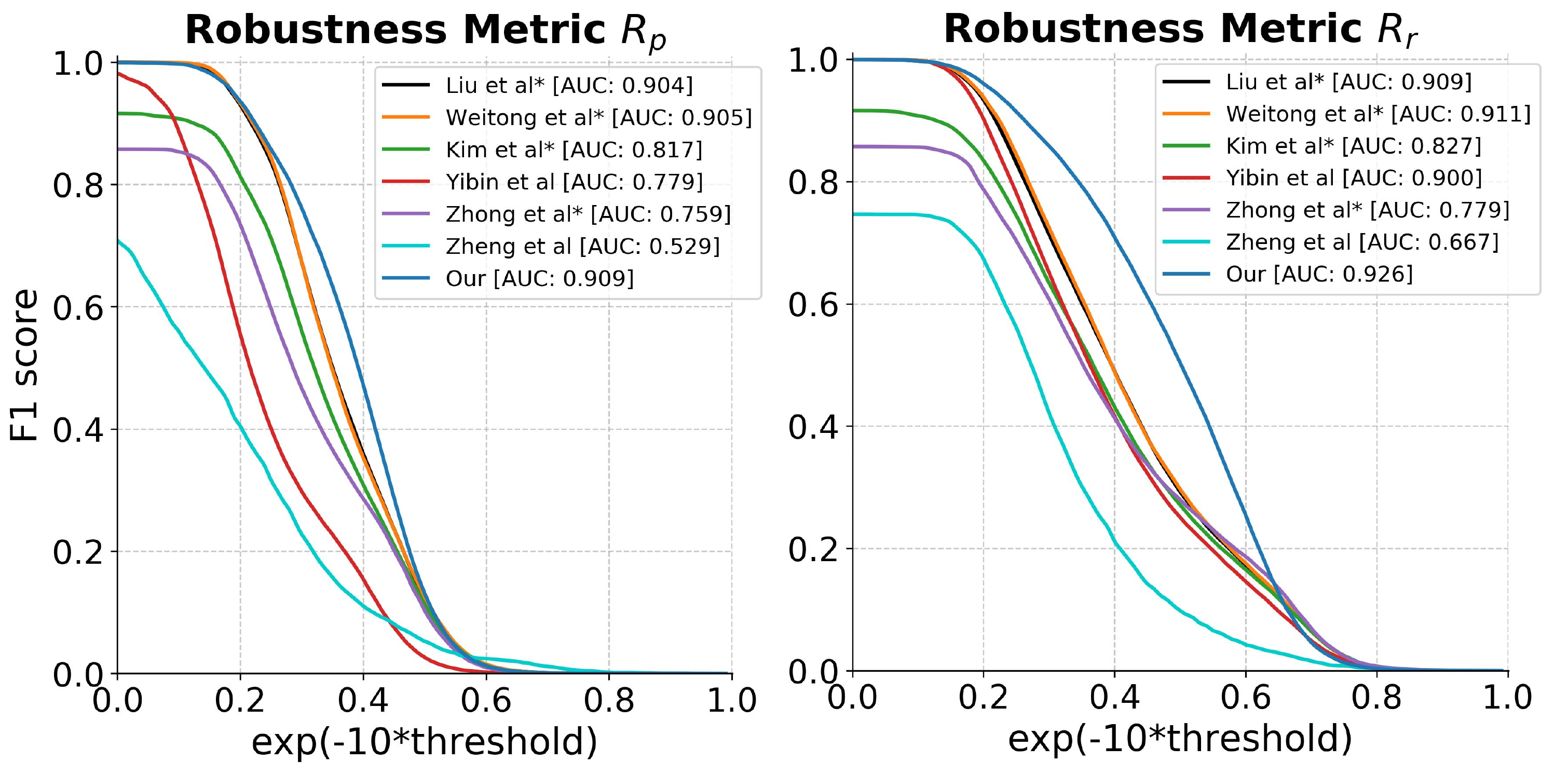}
  \caption{\textbf{Robustness Metrics.} $R_p$ and $R_r$ curves are presented for all teams competing in the SLAM challenge\cite{Zhao2024CVPR}. The robust metric is indicated by a high F1 score across the threshold range $[0, 1]$ using exponential mapping $exp(-10T)$. Our method (blue curve) has the most robust performance while running online without employ any loop closure or post-processing techniques.}
  \label{fig:robustness}
\end{figure}

An ideal evaluation metric should consider both accuracy and completeness \cite{Zhao2024CVPR}. While ATE effectively assesses trajectory accuracy, it falls short in capturing completeness. To address this limitation, we employ \textit{robustness metrics}\cite{Zhao2024CVPR} to consider both aspects. Tables~\ref{tab:RobustLidar} presents a comprehensive report of these metrics across various scenarios and algorithms. To see the value of these metrics, consider, e.g., that while the Floor01 ATE result for Zheng et al.'s method outperforms that of Corridor01, its $R_p$ value is worse due to the incomplete trajectory run. A summary of Robustness metrics across all other methods is shown in Fig. \ref{fig:robustness}. The blue curve, representing the $R_p$ and $R_r$ of our odometry method, is the highest across almost the entire range. 

\vspace{-0.15cm}
\subsection{Runtime Performance}
Our proposed method achieves a processing rate of 22 frames per second (FPS), with each frame taking 45 ms to process on an AMD Ryzen 7 3700X 8-Core CPU.
\section{Conclusion}
In this work, we proposed a map-based LiDAR localization that can anticipate and mitigate localization failures before the optimization process. Also, it provides a reliable observability estimation without heuristic threshold adjustment for different environments. Compared to existing methods, it achieves a remarkable 54\% improvement in accuracy over the second-best algorithms. In future work, we plan to incorporate observability into global scan registration to achieve better re-localization for the kidnapping problem.  


\balance

%

\bibliographystyle{IEEEtran}
\bibliography{root}
\end{document}